\documentclass{article}
\usepackage{ijcai20}

\usepackage{authblk}
\usepackage{times}
\usepackage{soul}
\usepackage{url}
\usepackage[hidelinks]{hyperref}
\usepackage[utf8]{inputenc}
\usepackage[small]{caption}
\usepackage{graphicx}
\usepackage{amsmath}
\usepackage{amsthm}
\usepackage{booktabs}
\usepackage{algorithm}
\usepackage{algorithmic}
\usepackage{multirow}
\usepackage{xcolor}
\usepackage{amsfonts}

\newcommand{\etal}{\textit{et al}.}
\newcommand{\ie}{\textit{i}.\textit{e}.}
\newcommand{\eg}{\textit{e}.\textit{g}.}

\title{Look Closer to Ground Better: Weakly-Supervised Temporal Grounding of Sentence in Video}

\author[Chen \etal]{
Zhenfang Chen$^1$\thanks{\quad Work done while Zhenfang Chen was a Research Intern with Tencent AI Lab.}\quad
Lin Ma$^2$\thanks{\quad Corresponding authors.}\quad Wenhan Luo$^2$ \quad Peng Tang$^3$ \quad Kwan-Yee K. Wong$^1$ \\
$^1$The University of Hong Kong  \quad $^2$Tencent AI Lab $^3$Huazhong University of Science and Technology \\
\texttt{\{zfchen, kykwong\}@cs.hku.hk} \\
\texttt{\{forest.linma, whluo.china\}@gmail.com} \\
\texttt{pengtang@hust.edu.cn}  
}

\begin{document}
\maketitle

\begin{abstract}
In this paper, we study the problem of weakly-supervised temporal grounding of sentence in video. Specifically, given an untrimmed video and a query sentence, our goal is to localize a temporal segment in the video that semantically corresponds to the query sentence, with no reliance on any temporal annotation during training. We propose a two-stage model to tackle this problem in a coarse-to-fine manner. In the coarse stage, we first generate a set of fixed-length temporal proposals using multi-scale sliding windows, and match their visual features against the sentence features to identify the best-matched proposal as a coarse grounding result. In the fine stage, we perform a fine-grained matching between the visual features of the frames in the best-matched proposal and the sentence features to locate the precise frame boundary of the fine grounding result.
Comprehensive experiments on the ActivityNet Captions dataset and the Charades-STA dataset demonstrate that our two-stage model achieves compelling performance. 
\end{abstract}

\vspace{-1.5em}
\section{Introduction}
Given a natural sentence and an untrimmed video, temporal video grounding~\cite{gao2017tall,hendricks17iccv} aims to determine the start and end timestamps of one segment in the video that semantically corresponds to the given sentence. 
Recently, much research effort \cite{aaai19:actionlocalization,xu2019multilevel,Liu_2018_ECCV,chen2018temporally,zhang2018man,he2019read} has been put into this research topic and great progress has been achieved. However, these existing methods focus on fully-supervised learning for temporal video grounding. Their successes heavily depend on the availability of fine-grained temporal annotations which are extremely labor-intensive and time-consuming to obtain. Moreover, as pointed out by~\cite{wang2017untrimmednets}, temporal boundaries for actions in videos are usually more subjective and inconsistent across different annotators than object boundaries in images. This makes the task of collecting temporal annotations difficult and non-trivial.

\begin{figure}[t]
    \centering
    \includegraphics[width=\linewidth]{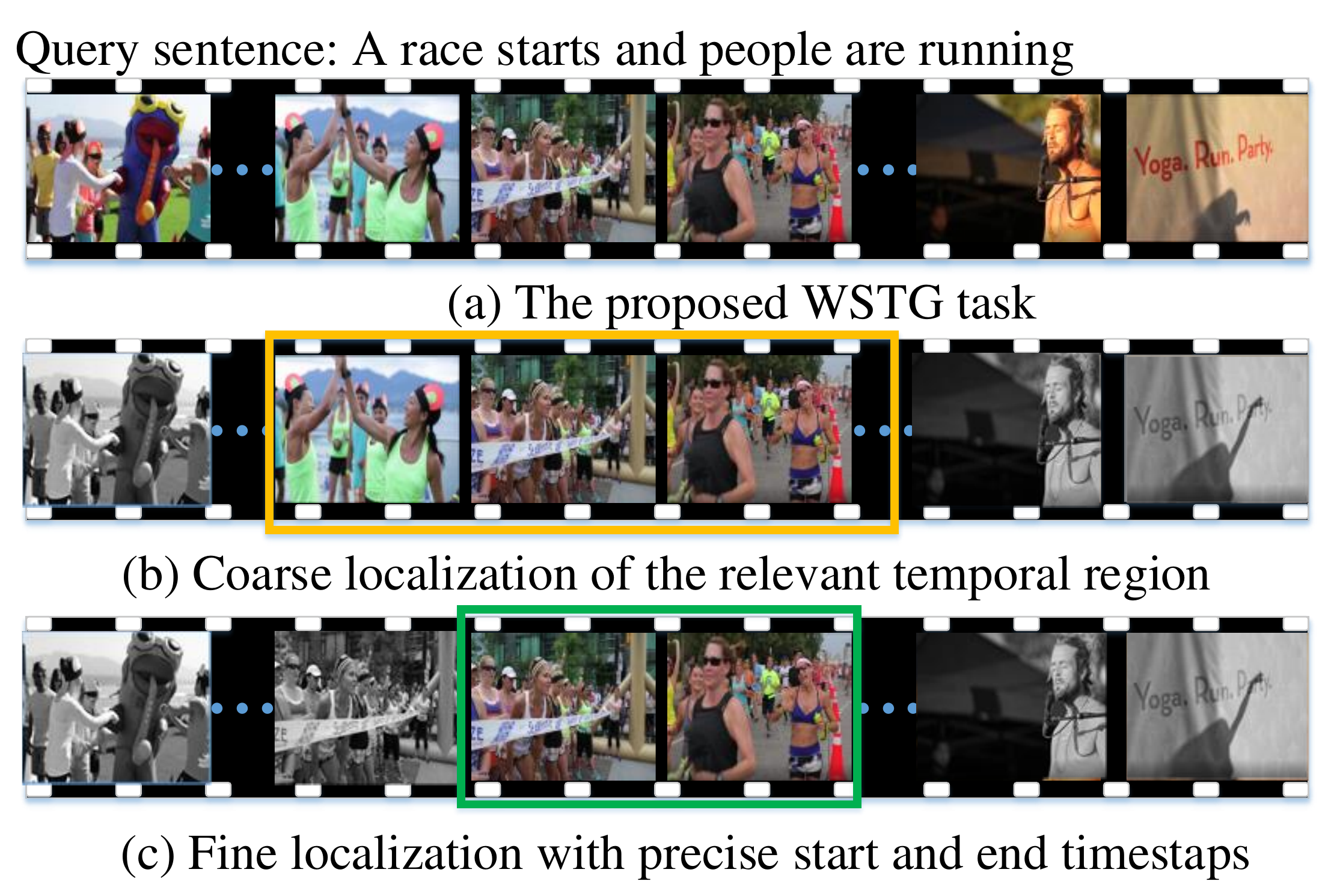}
    \vspace{-2em}
    \caption{(a) WSTG aims to localize a temporal segment in the video that semantically corresponds to a given sentence, with no reliance on any temporal annotations during training. 
    (b) A sliding window strategy is used to generate a set of fixed-length temporal proposals. Their visual features are matched against the sentence features to identify the best-matched proposal as a coarse grounding result. 
    (c) A fine-grained matching between the visual features of the frames in the best-matched proposal and the sentence features is carried out to locate the precise frame boundary of the fine grounding result. 
    }
    \label{fig:task}
    \vspace{-1.5em}
\end{figure}

In this paper, we study the problem of weakly-supervised temporal grounding of sentence in video (WSTG). As illustrated in Fig.~\ref{fig:task}(a), given an untrimmed video and a query sentence, WSTG aims to localize a temporal segment in the video that semantically corresponds to the sentence, with no reliance on any temporal annotation during training. By getting rid of the labor-intensive temporal annotations, WSTG requires much less effort in annotations and can easily scale to large datasets. Meanwhile, WSTG faces the challenge of how to precisely determine the start and end timestamps of the target segment in the video in the absence of precise temporal annotations as the supervision signal. 
Previous works like \cite{mithun2019weakly} usually handled the WSTG task by projecting features of two modalities into a common space and select a proposal from a set of pre-defined sliding windows. However, such methods did not consider the fine-grained interactions between the visual frames and the sentence, resulting in inferior performances.

Before we derive a solution to WSTG, let us first see how human beings carry out the temporal video grounding task. 
Typically, we first read and understand the query sentence at the beginning. With the semantic meaning of the sentence in mind, we then scan through the whole video and make a coarse localization of the relevant temporal region that roughly contains the target video segment (see Fig.~\ref{fig:task}(b)). Finally, we examine the frames in the localized coarse temporal region and determine the precise start and end timestamps of the target video segment (see Fig.~\ref{fig:task}(c)).

Based on the above observation, we propose a two-stage model to handle the WSTG task in a coarse-to-fine manner. Specifically, we use a bidirectional long short-term memory (Bi-LSTM) network to encode the sentence and capture its semantic meaning. Similarly, we use another Bi-LSTM network to encode and aggregate the contextual information of the sequential video frame representations. We then generate a set of fixed-length temporal proposals from the video using a sliding window strategy, and mine their matching relationships with the sentence. The proposal with the highest matching score is deemed to be the most relevant temporal region containing the target video segment. 
Note that, due to the simple strategy adopted for proposal generation, the start and end timestamps of the best-matched proposal often do not correspond to the precise start and end timestamps of the target video segment. Hence, the best-matched proposal provides only a coarse grounding result.
In order to achieve a more precise grounding result, we {\em look closer} at the coarse grounding result. Specifically, we mine the matching relationships between the sentence and each frame in the best-matched proposal and obtain their fine-grained semantic matching scores. The fine-grained matching scores are then grouped to generate the fine grounding result with precise start and end timestamps.


The contributions of this paper are summarized as follows.
    We propose a novel two-stage model to handle the WSTG task in a coarse-to-fine manner. Specifically, the coarse stage produces a coarse localization of the relevant region by mining the matching relationships between the sentence and the temporal proposals generated by a sliding window strategy. The fine stage exploits the relationships between the sentence and each frame in the coarse localization result, and generates an accurate grounding result with precise start and end timestamps. 
    Extensive experiments on ActivityNet Captions dataset and the Charades-STA dataset are conducted and analyzed, showing the effectiveness of the proposed model. 

\vspace{-0.5em}
\section{Related Work}
\vspace{-0.5em}
\noindent\textbf{Image Grounding.}
Grounding in images aims to localize a spatial region in the image that semantically matches the given sentence~\cite{hu2016natural,wang2016structured}. Recently, it has become a popular research topic.
 \cite{wang2016structured} proposed a structured matching between expressions and image regions. Recently, \cite{yu2018mattnet} and \cite{zhang2018grounding} considered using modular components and variational context to improve the performance. 
 
Weakly-supervised spatial grounding aims to localize a spatial region in the image which semantically matches the sentence, using only aligned image-sentence pair during training \cite{karpathy2015deep,rohrbach2016grounding}. \cite{karpathy2015deep} grounded phrases in images by mapping features of two modalities into a common space. \cite{rohrbach2016grounding} localized a region by reconstructing the query sentence. Later, weakly-supervised spatial grounding is introduced to the video domain \cite{huang2018finding,zhou2018weakly}. \cite{huang2018finding} localized a spatial region by modeling the reference relationships between video segments and instruction descriptions. \cite{zhou2018weakly} utilized the object interaction and loss weighting to localize a more precise spatial region. However, all these weakly-supervised spatial grounding methods still focused on spatial localization only. 

\noindent\textbf{Temporal Grounding.}
Temporally grounding a natural sentence in a video was first studied in \cite{hendricks17iccv,gao2017tall}, which aimed to localize a segment in the video that semantically corresponds to the given sentence. Since then, much effort has been made in this area \cite{xu2019multilevel,aaai19:actionlocalization}. \cite{aaai19:actionlocalization} generated discriminative proposals by integrating the semantic information of sentence queries into the proposal generation. \cite{xu2019multilevel} improved the performance by considering multilevel language and video features. \cite{Liu_2018_ECCV} explicitly modeled compositional reasoning by temporal modular networks. \cite{yuan2018find} introduced attention mechanisms to capture the fine-grained interaction between the sentence and the video. \cite{he2019read} solved the problem of temporal grounding by a policy learning strategy. \cite{zhang2018man} constructed a structured graph to model moment-wise temporal relationship. Although great progress has been made by these methods, they still relied on annotations of precise temporal boundaries of the target segment in the video, which are time-consuming and labor-intensive to obtain.

Recently, \cite{mithun2019weakly,gao2019wslln,duan2018weakly} focused on weakly-supervised temporal grounding. They usually selected a segment from a set of pre-defined proposals. The fine-grained interactions between visual frames and the sentence were not fully exploited, leading to inferior performance.

\vspace{-0.5em}
\section{Method}
\begin{figure}[t]
    \centering
    \includegraphics[width=\linewidth]{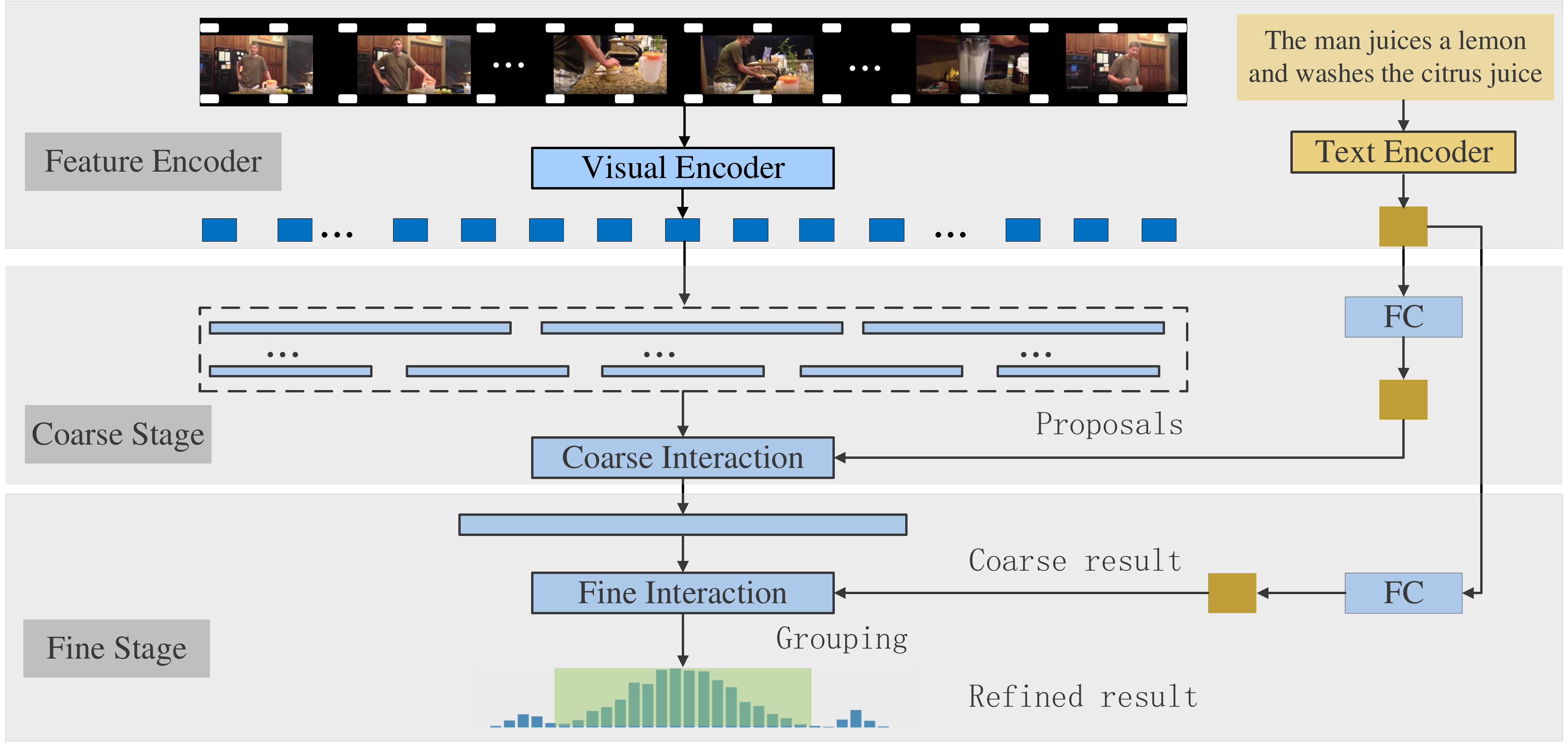}
    \vspace{-1em}
    \caption{The architecture of the proposed coarse-to-fine model. It contains three modules: feature encoder, coarse stage and fine stage. The feature encoder contains a visual encoder for extracting visual features from video clips and a text encoder for extracting textual features from the sentence. The coarse stage aims to select a coarse segment from the sliding window proposals. The fine stage targets at predicting the fine-grained matching scores between every frame in the coarse segment and the sentence. The final grounding result is generated by grouping the frames within the coarse segment.}
    \label{fig:frm}
    \vspace{-1.5em}
\end{figure}
Let $S=\{w_n\}_{n=1}^N$ denote a natural query sentence, where $w_n$ represents the $n$-th word and $N$ is the total number of words in $S$. Let $V = \{v_{t}\}_{t = 1}^{T}$\footnote{We extract one visual feature per second, which we call it a ``frame'' in this paper for simplicity.} denote an untrimmed video, where $v_t$ represents the $t$-th frame and $T$ is the total number of frames in $V$. Given $S$ and $V$, the proposed WSTG aims to localize a temporal segment $\phi=(t_s,t_e)$ in $V$ which expresses the same semantic meaning as $S$, where $t_s$ and $t_e$ denote respectively the start and end timestamps of $\phi$. Note that under the weakly-supervised setting, we do \textbf{NOT} have temporal boundary annotations during the training stage. 

In this paper, we propose a two-stage model to handle the WSTG task in a coarse-to-fine manner (see Fig.~\ref{fig:frm}). In the coarse stage, we target at localizing a coarse temporal region containing the target video segment, which is refined in the fine stage to determine the precise start and end timestamps of the target segment. Before the coarse stage, we first employ a text encoder to encode the semantic meaning of the sentence, and a visual encoder to encode and aggregate the contextual information of the video sequence (see Sec.~\ref{sec:feat_encoder}). Details of the coarse stage and the fine stage are given in Sec.~\ref{method:coarse} and Sec.~\ref{method:refine}, respectively, followed by details of the training and testing of our model in Sec.~\ref{train_test}.


\subsection{Feature Encoder}
\label{sec:feat_encoder}
In this section, we describe the details of our visual encoder and text encoder for extracting semantic features from the video and sentence respectively.

\paragraph{Visual Encoder.}
Given an untrimmed video $V=\{v_t\}_{t=1}^T$, we first extract a set of visual convolutional features $\textbf{C}=\{\textbf{c}_t\}_{t=1}^T$~(\eg, C3D features~\cite{tran2015learning}). Since convolutional features usually consider video characteristics within a short range only, we incorporate contextual information within a long range and cast aside irrelevant information using a Bi-LSTM to yield the corresponding hidden states $\textbf{H}_v=\{\textbf{h}_t^v\}_{t=1}^T$. For simplicity, we first project the convolutional features $\textbf{C}=\{\textbf{c}_t\}_{t=1}^T$ into a desired dimension (\ie, $512$ in our implementation) by a fully-connected (FC) layer to get  $\textbf{X}=\{\textbf{x}_t\}_{t=1}^T$. Formally, the above processes can be summarized as:
\vspace{-0.5em}
\begin{equation}
    \small 
    \label{eq:visen}
    \begin{aligned}
    \overrightarrow{\textbf{h}}_t^v &= \overrightarrow{\text{LSTM}_v}(\textbf{x}_t,~\overrightarrow{\textbf{h}}_{t-1}^v), \\
    \overleftarrow{\textbf{h}}_t^v &= \overleftarrow{\text{LSTM}_v}(\textbf{x}_t,~\overleftarrow{\textbf{h}}_{t+1}^v), \\
    \textbf{h}_t^v &=\overrightarrow{\textbf{h}_t^v}||\overleftarrow{\textbf{h}_t^v},
    \end{aligned}
    \vspace{-0.75em}
\end{equation}
where ``$||$'' denotes concatenation.

\paragraph{Text Encoder.}
We initialize each word $w_n$ in $S$ with the $300$-dimension glove vectors~\cite{pennington2014glove}. As such, we represent each sentence $S$ as a sequence of vectors $\{\textbf{w}_n\}_{n=1}^N$. Similarly, we encode these word sequences with another Bi-LSTM to exploit the semantic meaning expressed by the sentence and get a set of hidden states $\{\textbf{h}_n^s\}_{n=1}^N$. The first and last hidden states of the forward and backward LSTMs are concatenated to derive the representation for the whole sentence, \ie,~$\textbf{h}^s = \textbf{h}_N^s||\textbf{h}_0^s$.

\subsection{Coarse Stage}
\label{method:coarse}
Under the weakly-supervised setting, we do not have precise temporal annotations during training. We resort to multiple instance learning (MIL)~\cite{dietterich1997solving} to choose one temporal proposal from a set of fixed-length temporal proposals $P=\{p_k\}_{k=1}^K$, where $K$ is the total number of the proposals. As shown in Fig.~\ref{fig:frm}, the coarse stage consists of three main steps, namely temporal proposal generation, coarse interaction and coarse grounder.

\paragraph{Temporal Proposal Generation.}
Previous weakly-supervised image grounding methods \cite{karpathy2015deep,rohrbach2016grounding} usually obtain the grounding result by choosing an object proposal from a set of candidate proposals generated by some sophisticated proposal generation methods. To the best of our knowledge, the only semantic proposal generation method for actions described by natural language is \cite{aaai19:actionlocalization}. However, this method requires fine-grained temporal annotations, so it is not suitable for our WSTG task.

In this paper, we adopt a multi-scale sliding window strategy to generate fixed-length temporal proposals. We empirically choose window sizes $\{\frac{1}{6}l_v, \frac{1}{3}l_v, \frac{1}{2}l_v\}$, where $l_v$ is the average length of the videos in the training set. To generate the temporal proposals, we slide the windows across the timeline such that two consecutive proposals overlap with each other by 80\%. In this way, we generate a set of temporal proposals $P=\{p_k\}_{k=1}^K$, where $K$ denotes the total number of proposals generated from the video. For the $k$-th proposal, we represent it by $p_k=[t^k_s, t^k_e]$, with $t^k_s$ and $t^k_e$ denote the start and end timestamps of $p_k$ respectively.

\paragraph{Coarse Interaction.}
We first map each frame representation $\textbf{h}_t^v$ to $\hat{\textbf{h}}_t^v$ by a fully connected layer. We then generate a 
proposal feature $\textbf{f}^v_k$ for each proposal $p_k$ (which may have different temporal length depending on the sliding window size used) by max-pooling $\{\hat{\textbf{h}}_{t}^v\}_{t=t_s^k}^{t_e^k}$ along the temporal dimension. 
Similarly, we map the sentence representation $\textbf{h}^s$ to a sentence feature $\textbf{f}^s$ by another FC layer.

Given the proposal feature $\textbf{f}_k^v$ for the $k$-th proposal $p_k$ and the sentence feature $\textbf{f}^s$, a gating strategy is adopted to gate out the irrelevant parts and emphasize the relevant parts as follows:
\begin{equation}
    \small 
    \label{eq:gate}
    \begin{aligned}
        \textbf{g}_v=\sigma(\textbf{W}_v(\textbf{f}_k^v||\textbf{f}^s)+\textbf{b}_v), \qquad \hat{\textbf{f}}_k^v= \textbf{f}_k^v\odot \textbf{g}_v,  \\
        \textbf{g}_s=\sigma(\textbf{W}_s(\textbf{f}_k^v||\textbf{f}^s)+\textbf{b}_s), \qquad \hat{\textbf{f}}_k^s= \textbf{f}^s\odot \textbf{g}_s,  \\
    \end{aligned}
\end{equation}
where $\sigma$ is the sigmoid function, $\textbf{W}_v$, $\textbf{W}_s$, $\textbf{b}_v$, and $\textbf{b}_s$ are the learnable parameters. $\odot$ denotes element-wise dot product. 
As shown in Eq.~\ref{eq:match}, we interact the gated proposal feature $\hat{\textbf{f}}_k^v$ and the gated sentence feature $\hat{\textbf{f}}_k^s$  with a mechanism similar to ~\cite{gao2017tall}.
\begin{equation}
    \small 
    \label{eq:match}
    \textbf{f}_k^{v,s} =(\hat{\textbf{f}}_k^v + \hat{\textbf{f}_k^s})||(\hat{\textbf{f}}_k^v \odot \hat{\textbf{f}_k^s})||FC_1(\hat{\textbf{f}}_k^v||\hat{\textbf{f}_k^s}),
\end{equation}
where $FC_1$ is a FC layer.

\paragraph{Coarse Grounder.}
After interacting the features from both modalities, we use a coarse grounder to estimate the matching scores $M_c=\{m^c_k\}_{k=1}^K$ between the proposals and the sentence.
Inspired by the success of weakly-supervised detection in images and videos~\cite{bilen2016weakly,wang2017untrimmednets}, the coarse grounder contains two streams, classification stream and selection stream. The overall matching score for each proposal is summarized in Eq. \eqref{eq:match_coarse}.

\begin{equation}
    \small 
    \begin{aligned}
    \label{eq:match_coarse}
    m_k^c &= m_k^{cls} \cdot \ m_k^{slc} \\
         & = softmax_{cls}(FC_2(\textbf{f}_k^{v,s})) \cdot softmax_{slc}(FC_3(\textbf{f}_k^{v,s})).
    \end{aligned}
\end{equation}

\noindent{\textbf{Classification stream}} aims to classify whether a proposal matches the text description well and generate matching scores for each proposal independently. It uses a fully-connected layer ($FC_2$) to perform a binary classification for all the proposals and generate a similarity matrix $M_{cls} \in \mathbb{R}^{K \times 2}=\{m^{cls}_k\}_{k=1}^{K}$. A \textit{softmax} operator ($softmax_{cls}$) is adopted on the \textit{last} dimension to normalize the matching score for each proposal independently.

\noindent{\textbf{Selection stream}} aims to select a proposal from $P=\{p_k\}_{k=1}^K$ which matches the text description best and score regions relative to each other. Similar to classification stream, a fully connected layer ($FC_3$) is adopted to perform a binary classification. However, a \textit{softmax} operator ($softmax_{slc}$) is adopted on the \textit{first} dimension to generate matching scores $M_{slc} \in \mathbb{R}^{K \times 2}=\{m^{slc}_k\}_{k=1}^{K}$ to encourage competitions among the proposals.

The final matching score for each region is defined as the dot product of $m^{cls}_k$ and $m^{slc}_k$, enjoying the advantages of both streams.

As aforementioned, under the weakly-supervised setting, we resort to MIL to train our coarse stage model (see Sec.~\ref{train_test:train}). After training, given a set of generated proposals $P=\{p_k\}_{k=1}^K$ from a video, their matching scores $M_c=\{m(p_k, S)\}_{k=1}^K$ with the query sentence $S$ can be obtained by Eq.~\eqref{eq:match_coarse}. The proposal with the largest matching score is selected as the coarse grounding result, which we denote as $p^c=[t_s^c, t_e^c]$. Since the proposals are generated using a set of predefined sliding windows, it is difficult for $p^c$ to localize the target segment precisely.

\subsection{Fine Stage}
\label{method:refine}
To localize the target segment better, an additional fine stage is proposed to refine the coarse grounding result by adjusting the boundary of the best-matched proposal $p^c$. However, if we only consider boundary adjustment within $p^c$, we can only shrink the boundary. As such, to allow both expansion and shrinkage of the boundary, we extend the boundary of $p^c$ by a predefined scale. The expanded coarse grounding result is denoted as ${p'}^c=[{t'}_s^c,{t'}_e^c]$, with
\begin{equation} 
\small 
\begin{cases}
\ {t'}_s^c=\max(t_s^c - \lambda l_{\Delta}, 0), \\
\ {t'}_e^c=\min(t_e^c+\lambda l_{\Delta}, l'_v)
\end{cases}
\label{eq:expansion}
\end{equation}
where $l_{\Delta}$ is the length of $p^c$, $l'_v$ is the length of the video, and $\lambda$ is a scalar controlling the expansion.

\paragraph{Fine Interaction and Grounder.}
In the fine stage, we examine the fine-grained relationships between the given sentence and each frame in the expanded coarse grounding result ${p'}^c$.
Specifically, we first use two fully connected layers to map the $t$-th frame representation $\textbf{h}_t^v$ in $p'^c$ and the sentence representation $\textbf{h}^s$ to the frame feature ${\textbf{f}'}_t^v$ and sentence feature ${\textbf{f}'}^s$, respectively. We then adopt the same interaction strategy as in the coarse stage in Eq.~(\ref{eq:match}) to obtain the fine-grained interactive features. Moreover, we simply use a fully connected layer to generate the matching scores between the $t$-th frame and the text description, \ie ~$m'_t=FC_4(\textbf{f}_{t}^{v,s})$,  rather than the complicated two-stream grounder in the coarse stage.

Under the weakly-supervised setting, we again resort to MIL to train our fine stage model (see Sec.~\ref{train_test:train}). After training, we can densely predict the matching score between each frame in $p'^c$ and the sentence $S$. Based on these matching scores, we can localize the fine grounding result with precise start and end timestamps using a grouping strategy described below.

\paragraph{Grouping.}
\label{sec:grouping}
First, we normalize the matching scores between each frame and the sentence  
to $[0,1]$ by a linear transformation.
Similar to \cite{xiong2017pursuit}, we apply a watershed algorithm~\cite{roerdink2000the} to generate a set of refined proposals $\tilde{P}=\{\tilde{p}_i\}_{i=1}^I$ within the coarse grounding result ${p'^c}$, where $I$ is the total number of the refined proposals generated. And  the matching score $m_f$ for each proposal is obtained by aggregating the matching scores of each time step,
\ie~$m_f=\sum_{t={t_s^f}}^{t_e^f}{m'}_t$,
where $m'_t$ is the matching score between the $t$-{th} frame in the video and the query sentence $S$.

\subsection{Training}
\label{train_test}
\label{train_test:train}
For the WSTG task, since the temporal boundary annotations are unavailable, we cannot train the whole framework in a fully-supervised manner. Hence, we turn to MIL~\cite{dietterich1997solving} for training. The training of our model is performed in two stages. Specifically, we first train the feature encoder and the model in the coarse stage. Afterwards, with the feature encoder fixed, the model in the fine stage is trained with respect to the localized coarse results generated in the coarse stage.

\noindent{\textbf{Coarse Stage.}}
During training of the coarse stage, we estimate the matching score between the whole video and the sentence by accumulating matching scores of all the proposals, $m^c_{sum} \in \mathbb{R}^2=\sum_{k=1}^Km^c_k$ and selecting the proposal with the largest positive matching scores $m^c_{max} \in \mathbb{R}^2$. We optimize the model parameters with standard cross entropy:
\begin{equation}
    \small 
    \label{loss:c}
    \begin{aligned}
    \mathcal{L}_{c}=\sum_{j=0}^1(y[j]\log(m_{sum}^c[j]) + y[j]\log(m_{max}^c[j]))\,, \\
    \end{aligned}
\end{equation}
where $y$ is an one-hot vector. $y$ is set to $[0,1]$ if the video contains a segment that matches the given description well and to $[1,0]$ otherwise. This ranking loss will potentially encourage the aligned video segments in the aligned videos to generate higher matching scores than those of the misaligned videos.

\noindent{\textbf{Fine Stage.}}
After performing the training in the coarse stage, we can obtain the coarse grounding result and thereafter extend its boundary, denoted as  $p'^c=[t'^c_s, t'^c_e]$. In the fine stage, we simply use a fully connected layer to generate the matching score $m'_t$ between  the $t$-th frame and the given sentence $S$. We can thereby define the matching score between the query sentence  $S$ and the aligned video $V$ as:
\begin{equation} 
    \small 
    m'(V,S)=\max(m'_t),~t=t'^c_s,...,t'^c_e.
\end{equation}
We define the ranking loss $\mathcal{L}_{f}$ in the fine stage as:
\begin{equation}
    \small 
    \label{loss:f}
    \begin{aligned}
        \mathcal{L}_{f}=\sum_{S\neq~S'}&\sum_{V\neq~V'}[\max(m'(V',S)-m'(V,S)+\Delta,~0) \\
        +&\max(m'(V,S')-m'(V,S)+\Delta,~0)],
    \end{aligned}
\end{equation}
where $\Delta$ is a margin and is also set to $1$. This ranking loss in the fine stage will encourage the semantically correlated frames in the coarse result to produce higher matching scores than those of the uncorrelated ones.
\vspace{-0.5em}
\subsection{Inference}
\label{train_test:test}
During inference, we first extract the visual and textual features from the video and the sentence by the feature encoder. We then run the model in the coarse stage to obtain the matching scores for all the sliding window segments in the video and choose the video segment with the highest positive matching score as the coarse grounding result. Afterwards, the model in the fine stage is performed to derive the fine-grained matching scores for each frame in the expanded coarse grounding result. Based on the obtained fine-grained matching scores, we perform a grouping strategy, as introduced in Sec.~\ref{sec:grouping} to yield the grounding result with precise start and end timestamps.

\vspace{-0.5em}
\section{Experiments}
In this section, we evaluate the proposed method on two popular datasets, namely ActivityNet Captions and Charades-STA.
We first introduce the datasets, evaluation metrics and implementation details.
We then compare the proposed method with other methods. Finally, an ablation study is conducted to evaluate the effectiveness of each component. 

\vspace{-0.5em}
\subsection{Datasets}
\label{sec:datasets}
In this subsection, we introduce the datasets used to evaluate the WSTG task. Currently, there are four publicly available datasets for temporal grounding, namely DiDeMo~\cite{hendricks17iccv}, TACoS~\cite{regneri2013grounding}, ActivityNet Captions~\cite{krishna2017dense}, and Charades-STA~\cite{gao2017tall}. TACoS is unsuitable for the WSTG task. It contains only $127$ untrimmed videos for training and testing, which we suspect being insufficient to train a model for the WSTG task. DiDeMo is also unsuitable, because it only contains six segments for each video, which is too short to evaluate the performance of our algorithm. Thus, we conduct the experiments on ActivityNet Captions and Charades-STA.

\noindent{\textbf{ActivityNet Captions.}~\cite{krishna2017dense}.} ActivityNet Captions contains about $20$K videos with $100$K descriptions, which is the largest temporal grounding dataset. Following previous fully-supervised methods like~\cite{he2019read,xu2019multilevel}, we train on the training set and test on the validation sets since the caption annotations in the testing set are unavailable.

\noindent{\textbf{Charades-STA.}~\cite{gao2017tall}.} Charades-STA contains $9,848$ videos with $16,128$ clip-sentence pairs. Following the same split of previous fully-supervised methods like~\cite{gao2017tall}, $12,408$ and $3,720$ clip-sentence pairs are used for training and testing, respectively.

\vspace{-0.5em}
\subsection{Experimental Settings}
\label{sec:exp_setting}
\noindent{\textbf{Evaluation Metrics.}}
Following previous fully-supervised methods,~\eg,~\cite{gao2017tall,xu2019multilevel}, we adopt the metric ``R@1, IoU=$\eta$'' to evaluate the performance of temporal grounding. $\eta$ is set as $\{0.1, 0.3, 0.5\}$ for ActivityNet Captions and $\{0.3, 0.5, 0.7\}$ for Charades-STA, respectively. To evaluate the performance extensively, mIoU~(\ie, the mean IoU between the grounding results and ground-truths for all the sentence queries) is also reported.

\noindent\textbf{Implementation Details.} The video feature is usually generated with a temporal resolution. We extract one visual feature per second, which we regard as one ``frame'' in this paper. For fair comparisons, the corresponding visual feature is generated by C3D~\cite{tran2015learning} network, which is widely used in the supervised methods like \cite{gao2017tall,he2019read,xu2019multilevel}. The $300$-D glove~\cite{pennington2014glove} vector is used to initialize the word embeddings in the sentences. The size of hidden states of both Bi-LSTMs and all the FC layers are set to $512$. 

The length of the sliding windows in the coarse stage is set to $\{\frac{1}{6}, \frac{1}{3}, \frac{1}{2}\}$ of the average length of the videos in the training set, ~\ie, $\{20, 40, 60\}$ seconds for ActivityNet Captions and $\{5, 10, 15\}$ seconds for Charades-STA, respectively. The two consecutive windows have an overlap of $80\%$. The expand rates $\lambda$ in Eq.~(\ref{eq:expansion}) for ActivityNet Captions and Charades-STA are set to $1.8$ and $0.5$ by cross validation, respectively. The batch size for ActivityNet Captions and Charades-STA is set to $16$ and $32$. Specifically, we feed $16$ aligned video-sentence pairs to the network and construct the positive and negative pairs within a batch for efficiency. For the hyperparameters of the watershed algorithm in the grouping of the fine stage, we fix it the same as~\cite{xiong2017pursuit} for all the experiments to avoid
cumbersome hyperparameter tuning. We train the coarse stage for $100$ epochs and then train the fine stage for another $100$ epochs. Models are optimized by Adam algorithm with a learning rate of $10^{-3}$. 

\vspace{-0.5em}
\subsection{Comparisons with Other Methods}
\label{perf_com}
In this section, we compare our coarse-to-fine model with fully-supervised methods~\cite{gao2017tall,yuan2018find,xu2019multilevel,he2019read,aaai19:actionlocalization} and weakly-supervised \cite{mithun2019weakly,duan2018weakly,gao2019wslln}. Table~\ref{tab:act} and 
Table~\ref{tab:char} list the performance on ActivityNet Captions and Charades-STA, respectively. We notice that although our coarse-to-fine model employs only weak supervision of video-sentence pairs, it still achieves competitive performance on both datasets and even outperforms CTRL, a fully-supervised method. In ActivityNet Captions, while WSLLN (BERT)~\cite{gao2019wslln} has higher accuracy on ``R@1, IoU=0.1'', our method has the best performance on higher IoU thresholds and shows its capability to get more precise grounding results. It should be noted that WSLLN (BERT) relies on more powerful sentence encoder BERT~\cite{devlin2018bert}. When using a similar RNN sentence encoder like our model, WSLLN (GRU) achieves worse performance than our model in all metrics. In Charades-STA, our method outperforms the previous weakly-supervised method WSVMR~\cite{mithun2019weakly} in all thresholds. This indicates that our coarse-to-fine model can capture the semantic meaning of the video-sentence pairs and can temporally ground the sentence in the video. It demonstrates the effectiveness of our model on handling the WSTG task.

\begin{table}[t]
\centering
\small
 \resizebox{1\linewidth}{!}{
\begin{tabular}{|l|cccc|}
\hline
\multirow{2}{*}{Methods}             & R@1     &  R@1      &    R@1       &\multirow{2}{*}{mIoU} \\ 
                                     & IoU=0.1 &  IoU=0.3  &   IoU=0.5    &                   \\ 
\hline                                                                         
\hline
\multicolumn{5}{|c|}{fully-supervised methods}                             \\           
\hline
\hline  
CTRL~\cite{gao2017tall}              & 49.1    & 28.7      & 14.0         &     20.5    \\      
\cite{yuan2018find}             & \textbf{73.3}       & \textbf{55.7}   &    36.8  &  \textbf{37.0}\\  
\cite{xu2019multilevel}     & -       & 45.3      & 27.7   &           -         \\  
\cite{he2019read}           & -       & -         & \textbf{36.9}  &  - \\

\hline
\hline                                                                                        
\multicolumn{5}{|c|}{weakly-supervised methods}                             \\           
\hline 
\hline 
\cite{duan2018weakly}                             & 62.7        & 42.0       &  23.3 &  28.2\\ 

\cite{gao2019wslln}(GRU)                          & 74.0        & 42.3       &  22.5  & 31.8       \\  
\cite{gao2019wslln}(BERT)                         & \textbf{75.4}        & 42.8       &  22.7  & \textbf{32.2}       \\  
Ours                                              &  74.2       & \textbf{44.3}       &  \textbf{23.6} &  \textbf{32.2}  \\
\hline
\end{tabular}
}
\vspace{-1em}
\caption{Performance comparisons on ActivityNet Captions dataset.}
\label{tab:act}
\end{table}

\begin{table}[t]
\centering
\setlength{\tabcolsep}{0.3em}
\small
\begin{tabular}{|l|cccc|}
\hline
\multirow{2}{*}{Methods}              &  R@1      &    R@1    & R@1     & \multirow{2}{*}{mIoU}  \\ 
                                      &  IoU=0.3  &   IoU=0.5 & IoU=0.7 &             \\ 
\hline                                                                         
\hline
\multicolumn{5}{|c|}{fully-supervised methods}                             \\           
\hline
\hline 
CTRL~\cite{gao2017tall}                     & -         & 23.6   & 8.9   & - \\      
\cite{xu2019multilevel}            & ~\textbf{54.7}      & 35.6   & ~\textbf{15.8}  & - \\  
\cite{he2019read}                  & -         &~\textbf{36.7}   & -     & -  \\
SAP~\cite{aaai19:actionlocalization}        & -         & 27.4   & 12.6  & - \\
\hline
\hline                                                                                        
\multicolumn{5}{|c|}{weakly-supervised methods}    \\
\hline
\hline
\cite{mithun2019weakly}                                &  32.1   &    19.9    & 8.8        &    -     \\
Ours                                 &    \textbf{39.8} &       \textbf{27.3}   &       \textbf{12.9} &     \textbf{27.3}  \\
\hline
\end{tabular}
\vspace{-1em}
\caption{Performance comparisons on Charades-STA dataset.}
\label{tab:char}
\vspace{-1em}
\end{table}

\subsection{Ablation Study}
\label{abl_stu}
In this section, we first carry out ablation studies to investigate how the two stages and different grounders affect the performance.

\begin{table}[t]
\centering
\small
\begin{tabular}{|l|cccc|}
\hline
\multirow{2}{*}{Methods}             & R@1     &  R@1      &    R@1       &\multirow{2}{*}{mIoU} \\ 
                                     & IoU=0.1 &  IoU=0.3  &   IoU=0.5    &                   \\ 

\hline 
\hline 
Random                               & 26.2       & 12.1       &  4.8      & 9.0  \\ 
Coarse                               & 59.6       & 36.9       &  19.8     & 25.3    \\  
Full                    &  \textbf{74.2}       & \textbf{44.3}       &  \textbf{23.6} &  \textbf{32.2}  \\
\hline
\end{tabular}
\vspace{-1em}
\caption{Ablation study on ActivityNet Captions dataset.}
\label{abs:act}
\vspace{-1em}
\end{table}

\begin{table}[t]
\centering
\small
\begin{tabular}{|l|cccc|}
\hline
\multirow{2}{*}{Methods}              &  R@1      &    R@1    & R@1     & \multirow{2}{*}{mIoU}  \\ 
                                      &  IoU=0.3  &   IoU=0.5 & IoU=0.7 &             \\ 

\hline 
\hline
Random                                &  20.0   &    9.1       & 2.6        &    12.8   \\ 
Coarse                                & 32.1    &   21.0       & 11.8    & 21.2     \\  
Ours                                 &    \textbf{39.8} &       \textbf{27.3}   &       \textbf{12.9} &     \textbf{27.3}  \\
\hline
\end{tabular}
\vspace{-1em}
\caption{Ablation study on Charades-STA dataset.}
\label{abs:char}
\vspace{-1em}
\end{table}

\begin{figure}[th]
    \centering
    {People are roller blading in between orange cones.}
    \includegraphics[width=0.95\linewidth,height=0.13\linewidth]{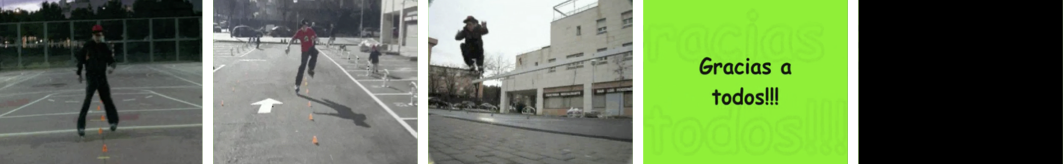}
    \includegraphics[width=0.95\linewidth,height=0.25\linewidth]{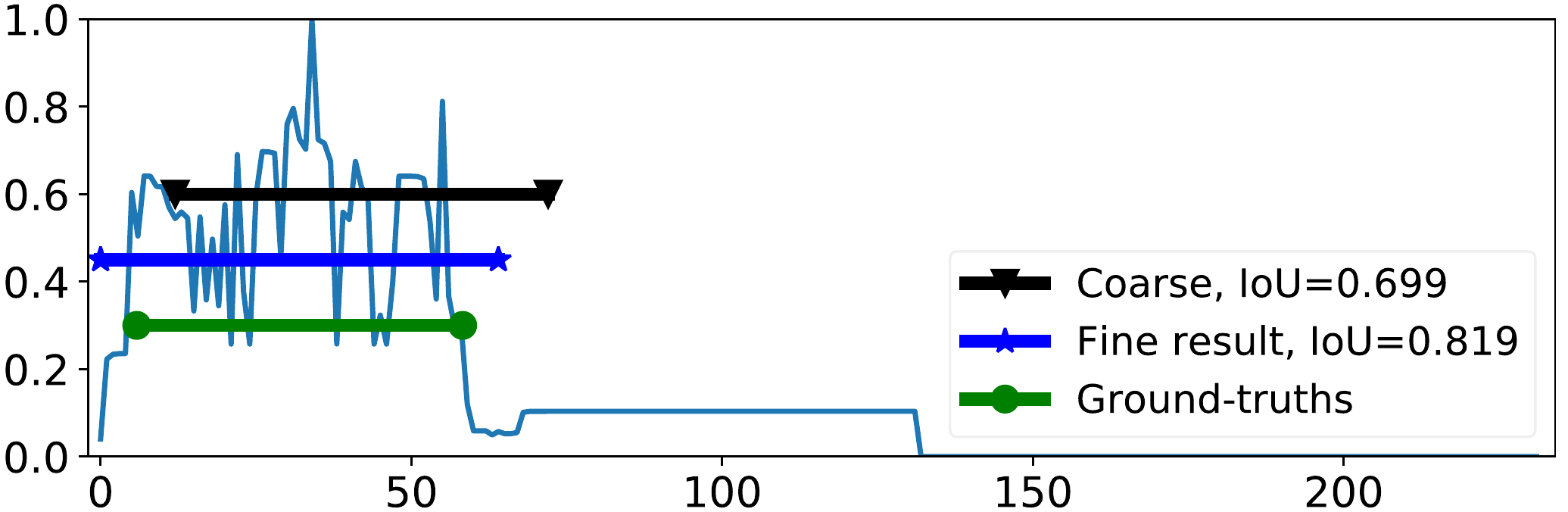}
    \vspace{-1em}
    \caption{A typical example of the grounding results. The fine-grained matching scores predicted by the \texttt{Full} model is shown with blue lines. The \texttt{Full} model provides more precise localization.}
    \label{fig:abs}
     \vspace{-1em}
\end{figure}

\noindent\textbf{{Effectiveness of the Coarse-to-Fine Model.}} We further show the effectiveness of the proposed method by ablation study. The baseline method that adopts the coarse stage only is termed as \texttt{Coarse}. Our full coarse-to-fine model is termed as \texttt{Full}. For better comparisons, we also show the performance by randomly choosing a segment from the sliding window proposals in \texttt{Coarse}.

Table~\ref{abs:act} and Table~\ref{abs:char} show the results, respectively. Based on the results, we have several observations. First, the \texttt{Coarse} model can effectively choose a segment from the sliding window proposals and is much better than randomly choosing a segment from the proposals. Second, our \texttt{Full} model is able to provide a fine-grained matching between each frame in the video and the query sentence and achieves better performance. As illustrated in Fig.\ref{fig:abs}, our model can provide more reliable grounding results by grouping the frames based on their fine-grained similarity with the query sentence and refines the boundary of the \texttt{Coarse} model. 

\noindent{\textbf{Effect of Different Grounders.}} We then show the effect of the different grounders in different stages. \texttt{FC} denotes that the grounder that simply use a fully connected layer to predict a matching score while \texttt{Two} represents the two-stream grounder in Eq.(~\ref{eq:match_coarse}).
Table~\ref{abs:grounder} shows the performance. On the coarse stage, we find that \texttt{Two} has better performance than \texttt{FC}, which we believe the reason is that the selection stream of two-stream grounder has encouraged the model to distinguish the segments within the same videos. We then fix the two-stream grounder in the coarse stage and compare the performance in the fine stage, we find that \texttt{FC} achieves better performance than \texttt{Two}. We suspect the reason is that usually multiple frames in the fine stage match the query sentence well. It may confuse the model to force a selection between similar frames through the selection branch.   

\begin{table}[th]
\centering
\setlength{\tabcolsep}{0.3em}
\small
\begin{tabular}{|l|cccc|}
\hline
\multirow{2}{*}{Methods}              &  R@1      &    R@1    & R@1     & \multirow{2}{*}{mIoU}  \\ 
                                      &  IoU=0.1  &   IoU=0.3 & IoU=0.5 &             \\ 

\hline 
\hline
Coarse (FC)                                &  59.7   & 35.1         & 17.3       &    24.2  \\
Coarse (Two)                               &  59.6   &  36.9        &  19.8      &  25.3    \\  
Coarse (Two) + Fine(Two)                   &  67.1   &   42.5       &  \textbf{24.1}      &  29.6     \\
Coarse (Two) + Fine(FC)                    &  \textbf{74.2}   & \textbf{44.3}       &  23.6 &  \textbf{32.2}  \\
\hline
\end{tabular}
\vspace{-1em}
\caption{Effect of different grounders on ActivityNet Caption.}
\vspace{-0.5em}
\label{abs:grounder}

\end{table}

\vspace{-2em}
\section{Conclusion}
\vspace{-0.5em}
In this paper, we tackle the WSTG task, which aims to temporally ground a given sentence in the video. During training, we pose no reliance on temporal annotations. To handle this task, we propose a novel coarse-to-fine model based on multiple instance learning. First, the coarse stage selects a rough segment from a set of predefined sliding windows, which semantically corresponds to the given sentence. Afterwards, the fine stage mines the fine-grained matching relationship between each frame in the coarse segment and the sentence. It thereby refines the boundary of the coarse segment by grouping the frames and get a more precise grounding result. Extensive experiments are conducted and analyzed, illustrating that the proposed model can achieve competitive performance on the widely-used ActivityNet Captions and Charades-STA datasets.   

\bibliographystyle{named}
\bibliography{egbib}

\begin{thebibliography}{}

\bibitem[\protect\citeauthoryear{Bilen and Vedaldi}{2016}]{bilen2016weakly}
Hakan Bilen and Andrea Vedaldi.
\newblock Weakly supervised deep detection networks.
\newblock In {\em CVPR}, 2016.

\bibitem[\protect\citeauthoryear{Chen and
  Jiang}{2019}]{aaai19:actionlocalization}
Shaoxiang Chen and Yu-Gang Jiang.
\newblock Semantic proposal for activity localization in videos via sentence
  query.
\newblock In {\em AAAI}, 2019.

\bibitem[\protect\citeauthoryear{Chen \bgroup \em et al.\egroup
  }{2018}]{chen2018temporally}
Jingyuan Chen, Xinpeng Chen, Lin Ma, Zequn Jie, and Tat-Seng Chua.
\newblock Temporally grounding natural sentence in video.
\newblock In {\em EMNLP}, 2018.

\bibitem[\protect\citeauthoryear{Devlin \bgroup \em et al.\egroup
  }{2018}]{devlin2018bert}
Jacob Devlin, Ming-Wei Chang, Kenton Lee, and Kristina Toutanova.
\newblock Bert: Pre-training of deep bidirectional transformers for language
  understanding.
\newblock {\em arXiv preprint arXiv:1810.04805}, 2018.

\bibitem[\protect\citeauthoryear{Dietterich \bgroup \em et al.\egroup
  }{1997}]{dietterich1997solving}
Thomas~G Dietterich, Richard~H Lathrop, and Tom{\'a}s Lozano-P{\'e}rez.
\newblock Solving the multiple instance problem with axis-parallel rectangles.
\newblock {\em Artificial intelligence}, 1997.

\bibitem[\protect\citeauthoryear{Duan \bgroup \em et al.\egroup
  }{2018}]{duan2018weakly}
Xuguang Duan, Wenbing Huang, Chuang Gan, Jingdong Wang, Wenwu Zhu, and Junzhou
  Huang.
\newblock Weakly supervised dense event captioning in videos.
\newblock In {\em NIPS}, 2018.

\bibitem[\protect\citeauthoryear{Gao \bgroup \em et al.\egroup
  }{2017}]{gao2017tall}
Jiyang Gao, Chen Sun, Zhenheng Yang, and Ram Nevatia.
\newblock Tall: Temporal activity localization via language query.
\newblock In {\em ICCV}, 2017.

\bibitem[\protect\citeauthoryear{Gao \bgroup \em et al.\egroup
  }{2019}]{gao2019wslln}
Mingfei Gao, Larry~S Davis, Richard Socher, and Caiming Xiong.
\newblock Wslln: Weakly supervised natural language localization networks.
\newblock {\em EMNLP}, 2019.

\bibitem[\protect\citeauthoryear{He \bgroup \em et al.\egroup
  }{2019}]{he2019read}
Dongliang He, Xiang Zhao, Jizhou Huang, Fu~Li, Xiao Liu, and Shilei Wen.
\newblock Read, watch, and move: Reinforcement learning for temporally
  grounding natural language descriptions in videos.
\newblock {\em AAAI}, 2019.

\bibitem[\protect\citeauthoryear{Hendricks \bgroup \em et al.\egroup
  }{2017}]{hendricks17iccv}
Lisa~Anne Hendricks, Oliver Wang, Eli Shechtman, Josef Sivic, Trevor Darrell,
  and Bryan Russell.
\newblock Localizing moments in video with natural language.
\newblock In {\em ICCV}, 2017.

\bibitem[\protect\citeauthoryear{Hu \bgroup \em et al.\egroup
  }{2016}]{hu2016natural}
Ronghang Hu, Huazhe Xu, Marcus Rohrbach, Jiashi Feng, Kate Saenko, and Trevor
  Darrell.
\newblock Natural language object retrieval.
\newblock In {\em CVPR}, 2016.

\bibitem[\protect\citeauthoryear{Huang \bgroup \em et al.\egroup
  }{2018}]{huang2018finding}
De-An Huang, Shyamal Buch, Lucio Dery, Animesh Garg, Li~Fei-Fei, and
  Juan~Carlos Niebles.
\newblock Finding “it”: Weakly-supervised reference-aware visual grounding
  in instructional videos.
\newblock CVPR, 2018.

\bibitem[\protect\citeauthoryear{Karpathy and Fei-Fei}{2015}]{karpathy2015deep}
Andrej Karpathy and Li~Fei-Fei.
\newblock Deep visual-semantic alignments for generating image descriptions.
\newblock In {\em CVPR}, 2015.

\bibitem[\protect\citeauthoryear{Krishna \bgroup \em et al.\egroup
  }{2017}]{krishna2017dense}
Ranjay Krishna, Kenji Hata, Frederic Ren, Li~Fei-Fei, and Juan Carlos~Niebles.
\newblock Dense-captioning events in videos.
\newblock In {\em ICCV}, 2017.

\bibitem[\protect\citeauthoryear{Liu \bgroup \em et al.\egroup
  }{2018}]{Liu_2018_ECCV}
Bingbin Liu, Serena Yeung, Edward Chou, De-An Huang, Li~Fei-Fei, and Juan
  Carlos~Niebles.
\newblock Temporal modular networks for retrieving complex compositional
  activities in videos.
\newblock In {\em ECCV}, September 2018.

\bibitem[\protect\citeauthoryear{Mithun \bgroup \em et al.\egroup
  }{2019}]{mithun2019weakly}
Niluthpol~Chowdhury Mithun, Sujoy Paul, and Amit~K Roy-Chowdhury.
\newblock Weakly supervised video moment retrieval from text queries.
\newblock In {\em CVPR}, 2019.

\bibitem[\protect\citeauthoryear{Pennington \bgroup \em et al.\egroup
  }{2014}]{pennington2014glove}
Jeffrey Pennington, Richard Socher, and Christopher Manning.
\newblock Glove: Global vectors for word representation.
\newblock In {\em EMNLP}, 2014.

\bibitem[\protect\citeauthoryear{Regneri \bgroup \em et al.\egroup
  }{2013}]{regneri2013grounding}
Michaela Regneri, Marcus Rohrbach, Dominikus Wetzel, Stefan Thater, Bernt
  Schiele, and Manfred Pinkal.
\newblock Grounding action descriptions in videos.
\newblock {\em TACL}, 2013.

\bibitem[\protect\citeauthoryear{Roerdink and Meijster}{2000}]{roerdink2000the}
Jos B T~M Roerdink and Arnold Meijster.
\newblock The watershed transform: definitions, algorithms and parallelization
  strategies.
\newblock {\em Fundamenta Informaticae}, 2000.

\bibitem[\protect\citeauthoryear{Rohrbach \bgroup \em et al.\egroup
  }{2016}]{rohrbach2016grounding}
Anna Rohrbach, Marcus Rohrbach, Ronghang Hu, Trevor Darrell, and Bernt Schiele.
\newblock Grounding of textual phrases in images by reconstruction.
\newblock In {\em ECCV}, 2016.

\bibitem[\protect\citeauthoryear{Tran \bgroup \em et al.\egroup
  }{2015}]{tran2015learning}
Du~Tran, Lubomir Bourdev, Rob Fergus, Lorenzo Torresani, and Manohar Paluri.
\newblock Learning spatiotemporal features with 3d convolutional networks.
\newblock In {\em ICCV}, 2015.

\bibitem[\protect\citeauthoryear{Wang \bgroup \em et al.\egroup
  }{2016}]{wang2016structured}
Mingzhe Wang, Mahmoud Azab, Noriyuki Kojima, Rada Mihalcea, and Jia Deng.
\newblock Structured matching for phrase localization.
\newblock In {\em ECCV}, 2016.

\bibitem[\protect\citeauthoryear{Wang \bgroup \em et al.\egroup
  }{2017}]{wang2017untrimmednets}
Limin Wang, Yuanjun Xiong, Dahua Lin, and Luc Van~Gool.
\newblock Untrimmednets for weakly supervised action recognition and detection.
\newblock In {\em CVPR}, 2017.

\bibitem[\protect\citeauthoryear{Xiong \bgroup \em et al.\egroup
  }{2017}]{xiong2017pursuit}
Yuanjun Xiong, Yue Zhao, Limin Wang, Dahua Lin, and Xiaoou Tang.
\newblock A pursuit of temporal accuracy in general activity detection.
\newblock {\em arXiv preprint arXiv:1703.02716}, 2017.

\bibitem[\protect\citeauthoryear{Xu \bgroup \em et al.\egroup
  }{2019}]{xu2019multilevel}
Huijuan Xu, Kun He, L~Sigal, S~Sclaroff, and K~Saenko.
\newblock Multilevel language and vision integration for text-to-clip
  retrieval.
\newblock In {\em AAAI}, 2019.

\bibitem[\protect\citeauthoryear{Yu \bgroup \em et al.\egroup
  }{2018}]{yu2018mattnet}
Licheng Yu, Zhe Lin, Xiaohui Shen, Jimei Yang, Xin Lu, Mohit Bansal, and
  Tamara~L Berg.
\newblock Mattnet: Modular attention network for referring expression
  comprehension.
\newblock In {\em CVPR}, 2018.

\bibitem[\protect\citeauthoryear{Yuan \bgroup \em et al.\egroup
  }{2019}]{yuan2018find}
Yitian Yuan, Tao Mei, and Wenwu Zhu.
\newblock To find where you talk: Temporal sentence localization in video with
  attention based location regression.
\newblock {\em AAAI}, 2019.

\bibitem[\protect\citeauthoryear{Zhang \bgroup \em et al.\egroup
  }{2018}]{zhang2018grounding}
Hanwang Zhang, Yulei Niu, and Shih-Fu Chang.
\newblock Grounding referring expressions in images by variational context.
\newblock In {\em CVPR}, 2018.

\bibitem[\protect\citeauthoryear{Zhang \bgroup \em et al.\egroup
  }{2019}]{zhang2018man}
Da~Zhang, Xiyang Dai, Xin Wang, Yuan-Fang Wang, and Larry~S Davis.
\newblock Man: Moment alignment network for natural language moment retrieval
  via iterative graph adjustment.
\newblock {\em CVPR}, 2019.

\bibitem[\protect\citeauthoryear{Zhou \bgroup \em et al.\egroup
  }{2018}]{zhou2018weakly}
Luowei Zhou, Nathan Louis, and Jason~J Corso.
\newblock Weakly-supervised video object grounding from text by loss weighting
  and object interaction.
\newblock {\em BMVC}, 2018.

\end{thebibliography}
\end{document}